\documentclass[11pt]{article}

\usepackage{dialogue}
\usepackage{graphicx}

  \usepackage{cmap}
  \usepackage[T2A,T1]{fontenc}
  \usepackage[utf8]{inputenc}
  \usepackage{times}
  \usepackage{latexsym}
  \usepackage{substitutefont}
  \substitutefont{T2A}{\familydefault}{cmr}

\usepackage[russian,british]{babel}
\usepackage{url}
\usepackage{pgf}
\usepackage{enumitem}
\usepackage{float}

\usepackage{covington} 

\usepackage{hyperref}

\newcommand{\russianexample}[2]{\foreignlanguage{russian}{‘#1’} ‘#2’}

\dialogfinalcopy 

\title{The Pilot Corpus of the English Semantic Sketches}

\author{Maria Petrova \\
  ABBYY \\ 
  Moscow, Russia  \\
  {\tt m.petrova@abbyy.com} \\\And
  Maria Ponomareva \\
  HSE, ABBYY \\
  Moscow, Russia \\
  {\tt maria.ponomareva@abbyy.com} \\\AND
  Alexandra Ivoylova \\
  RSUH, MIPT, ABBYY \\ 
  Moscow, Russia \\
  {\tt aleksandra.ivoilova@abbyy.com}}

\date{}

\begin{document}
\maketitle
\begin{abstract}

  The paper is devoted to the creation of the semantic sketches for English verbs. The pilot corpus consists of the English-Russian sketch pairs and is aimed to show what kind of contrastive studies the sketches help to conduct. Special attention is paid to the cross-language differences between the sketches with similar semantics. Moreover, we discuss the process of building a semantic sketch, and analyse the mistakes that could give insight to the linguistic nature of sketches.
  
  \textbf{Keywords:} word sketches, semantic sketches, frame semantics, word sense disambiguation, corpus lexicography
  
  \textbf{DOI:} 10.28995/2075-7182-2022-20-XX-XX
\end{abstract}

\selectlanguage{russian}
\begin{center}
  \russiantitle{Пилотный корпус английских семантических скетчей}

  \medskip \setlength\tabcolsep{1cm}
  \begin{tabular}{cc}
  
  \author
  \textbf{Мария Петрова}     & \textbf{Мария Пономарева}\\
      ABBYY & ВШЭ, ABBYY \\
       Москва, Россия & Москва, Россия \\
      {\tt m.petrova@abbyy.com} & {\tt maria.ponomareva@abbyy.com}\\
  \end{tabular}
  
  \begin{tabular}{c}
    \textbf{Александра Ивойлова}\\
      РГГУ, МФТИ, ABBYY\\
      Москва, Россия\\
      {\tt aleksandra.ivoilova@abbyy.com} \\
  \end{tabular}
  \medskip
\end{center}

\begin{abstract}
 Работа посвящена созданию семантических скетчей для глаголов английского языка. Пилотный корпус состоит из англо-русских пар скетчей, на примере которых демонстрируется, какие сопоставительные исследования скетчи позволяют проводить. 
 Особое внимание уделяется межъязыковым различиям скетчей одного семантического поля в разных языках.
 Кроме того, в статье обсуждается процесс построения скетча, возможные ошибки и их лингвистическая природа.
  
  \textbf{Ключевые слова:} скетчи слов, семантические скетчи, семантика фреймов, разрешение лексической многозначности, корпусная лексикография
\end{abstract}
\selectlanguage{british}

\section{Introduction}
\label{intro}

In the current paper, we present the pilot corpus of the English semantic sketches and compare the English sketches with their Russian counterparts. The semantic sketch is a lexicographical portrait of a verb, which is built on a large dataset of contexts and includes the most frequent dependencies of the verb. The sketches consist of the semantic roles which, in turn, are filled with the most typical representatives of the roles. 

The influence of context on word recognition has been well-known for quite a time. Semantic context allows faster word recognition and the inferring of the skipped words while reading. The research in this area has been conducted in psycholinguistics since the 1970s, with the earliest works by \cite{tweedy1977semantic} and \cite{becker1980semantic}. Here the focus is on visual word recognition while reading and word recognition by bilingual persons \cite{assche2012bilingual}. Another aspect of the topic is the automatic inferring of the skipped words by context, widely known as a common NLP task today. 

The ability to represent the word by its context is the central idea of distributional semantics. It serves as a basis for the bag-of-words task, which is a training objective for static vectors like word2vec \cite{Mikolov13} and FastText \cite{bojanowski2017enriching}. In the approach, the context has a set length, and the words entering the fixed window are considered equally. 

The semantic sketches do not have such disadvantages, as they are based on the result of the semantic parsing and therefore take into account not all the words occurring in the context, but only the words that semantically depend on the given core. That is, we take not the linearly nearest tokens, but the tokens close in the parsing graph, where the type of the links is considered as well.

The BERT \cite{devlin-etal-2019-bert} contextual embeddings, which followed the static vectors and became a state-of-the-art solution for meaning representation, also rely on the idea of expressing word semantics through its context, using the objective of masked language modeling.

One of the main weaknesses of all vector representations is their interpretation and quality evaluation. The common practice is to consider the vectors as good, if they allow one to get the necessary quality for the down-stream task.

The advantage of the semantic sketches is in their interpretability and clear creation process. The sketches can be regarded as human-interpretable representation of word meanings, which one gets automatically with the help of the statistical methods used on the large text datasets.

The semantic sketches were first demonstrated in \cite{Detkova2020}, where we presented the idea of the semantic sketches itself and analysed the semantic mark-up used for building the sketches. Further, the pilot corpus of the Russian sketches has been created \cite{semsketches2021}. Herein, we have continued the work and created the pilot corpus of the English semantic sketches. The corpus is bi-lingual: each English sketch is accompanied by the Russian analogue with the same semantics, so one can compare the English sketch with the Russian one and analyse the contrastive differences between the sketches. Thereby, the contribution of the current paper is the creation of the English semantic sketches, on the one hand, and the creation of the parallel bilingual sketch corpus -- on the other.

The structure of the paper is as follows. First, we briefly characterise the semantic sketches themselves. Second -- give a description of the suggested corpus and explain what kind of verbs it contains. After that, we analyze the mistakes one faces when building the sketches, and focus on the cross-language differences between the sketches with similar semantics. In conclusion, we summarize the results.

\section{Semantic Sketches}

The idea to represent word compatibility in the form of the sketch belongs to Adam Kilgarriff \cite{kilgarriff2014sketch} and is currently realized in the Sketch Engine project\footnote{www.sketchengine.eu}. Verbal dependencies are classified according to their syntactic roles and statistically ranged, which allows one to see all of the most frequent syntactic dependencies of the verb at the same time.

The problem is that the syntactic sketches do not differentiate between various meanings of the verbs and combine all possible meanings in one sketch. To overcome this problem, we suggested the semantic sketches, which take the semantic models into account and classify the dependencies by their semantic relations with the core instead of their surface realizations  \cite{Detkova2020}. For instance, see fig. \ref{fig:focus} with the sketch of the verb ‘to focus’ in the meaning 'to concentrate on smth., to pay special attention to smth.':

\begin{figure}[H]
\centering
\includegraphics[width=15cm]{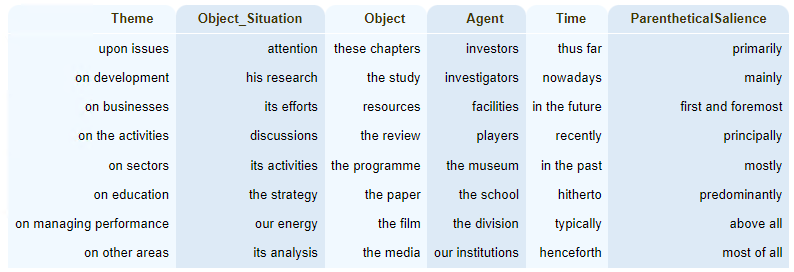}
\caption{The sketch for the verb ‘to focus:TO\_FOCUS’}
\label{fig:focus}
\end{figure}

Such sketches are built for each meaning separately, however, it demands a significant text corpus with full semantic mark-up. The authors settled on the Compreno mark-up built by the Compreno parser, which includes not only actant dependencies, but all possible links.

In the Compreno model, all words are presented in the form of a thesaurus-like semantic hierarchy, which consists of the semantic classes (semantic fields), and a set of the semantic roles for the classes (for detail, see \cite{anisimovich2012syntactic}, \cite{petrova2014compreno}). If a verb has several meanings, it enters several semantic classes with its own semantic model each. The semantic class is specified for each sketch.

\section{English SemSketches Corpus}

The SemSketches pilot corpus consists of 100 English sketches which are manually checked. It means that we have chosen the sketches manually according to their quality. The sketches are built on the corpus of the English texts comprising different genres, such as technical texts, news, fiction, and containing 14 million syntactic verbal links, that is, links which depend on the verbal cores. 

\begin{figure}[H]
\centering
\includegraphics[width=15cm]{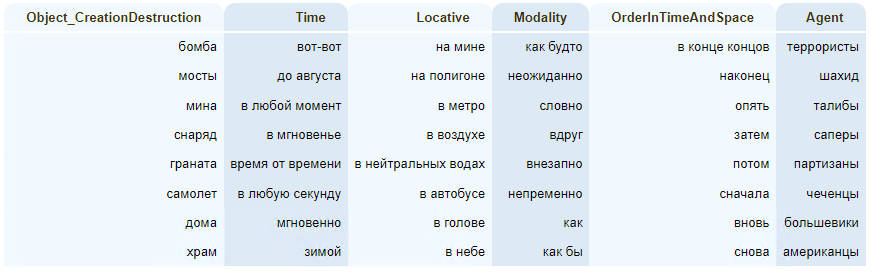}
\caption{The sketch for the verb ‘\foreignlanguage{russian}{взорвать}’:TO\_BLOW\_UP’}  
\label{fig:vzorvat}
\end{figure}

Each English sketch is provided with the parallel Russian sketch from the same semantic class, as shown in fig. \ref{fig:explode} and \ref{fig:vzorvat}:

\begin{figure}[H]
\centering
\includegraphics[width=15cm]{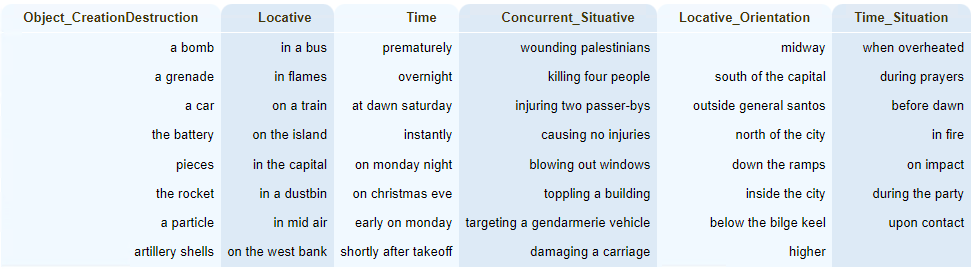}
\caption{The sketch for the verb ‘to explode:TO\_BLOW\_UP’}  
\label{fig:explode}
\end{figure}

For 100 English sketches, 84 Russian sketches are used: it means that some Russian sketches correspond to more than one English sketch. Totally, the corpus includes 113 English-Russian sketch pairs.

The choice of the English verbs is based on the Russian corpus which was built in \cite{semsketches2021}. The Russian corpus, in turn, includes only polysemantic verbs as an important point is to investigate how good the sketches can differentiate between various meanings of the verbs. 

To form the English sample, we have taken the verbs from the same semantic classes and set the threshold of 200 semantic links for each English verb: it means, the verb must have at least 200 links in the English texts corpus. (For comparison, the threshold for Russian verbs was 2000 links, but the Russian sketches were collected on the bigger dataset which includes more than 36 million links.)

After it, 100 English sketches were chosen, which met the above mentioned criteria and seemed to be enough representative to show the ability of the sketches to deal with polysemy, word sense disambiguation (WSD) problem, and asymmetrical compatibility of the verbs with similar semantics in different languages. Of course, the pilot corpus of 100 sketches is not enough for conducting representative contrastive research, however, certain observations seem to be of interest for comparative studies even on the small sample, as it is demonstrated below.

\section{What the mistakes in the sketches demonstrate}

The sketches are based on (1) the semantic relations the verb has in the text collection; (2) the work of the parser which classifies the relations according to their semantic roles and defines the meanings of the verbs. Therefore, the view of the sketch depends on the number of links the verb has in the corpus, and on the correctness of the parser’s work.
Herein the following mistakes are possible, which concern the automatic generation of the sketches.

\subsection{‘Empty’ sketches}

\begin{figure}[H]
\centering
\includegraphics[width=15cm]{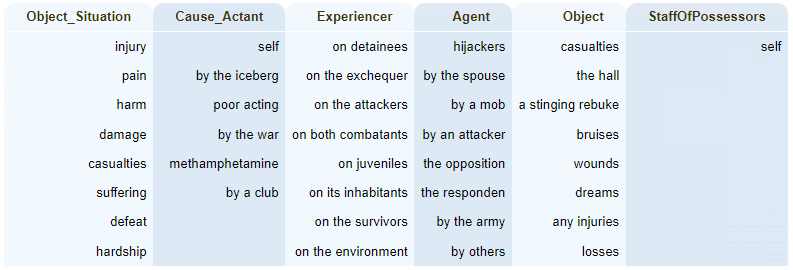}
\caption{The sketch for the verb ‘to inflict:TO\_BRING\_STATE\_TO\_SMB’}  
\label{fig:inflict}
\end{figure}

The insufficient number of links leads to partly ‘empty’ sketches, where the semantic roles contain very few fillers, up to only one. So when the semantic role column is partly empty, it can mean that the number of the role’s links in the corpus turned out to be poor (for instance, see the [Cause\_Actant] slot in the sketch for ‘inflict’ on fig. \ref{fig:inflict}]). As the number of texts grows, this problem occurs rarer. 

Another reason for the lack of fillers comes from the narrowness of the semantic role filling. That is, slots like [Object] or [Cause] have rather wide filling, while [Locative] and [Time] are more restricted in this respect. In turn, the Compreno parser has a large set of characteristic slots (for size, colour, speed, modality, and so on), so some slots possess rather narrow semantics and include a small set of fillers (like the [StaffOfPossessors] slot in the same sketch on fig. \ref{fig:inflict}).
 
Moreover, there are verbs with narrow compatibility, such as lexical functions. For instance, see the [Object] slot in the sketch of ‘\foreignlanguage{russian}{играть}:TO\_COMMIT’ (fig. \ref{fig:igrat}).

In sketches like these, empty lines in the semantic slots are correct.

\subsection{Incorrect semantic roles or incorrect fillers}

Other errors concern either the incorrect choice of the semantic slot for the given verb meaning, or the wrong fillers of the slot. As one of the key points is to examine how well the sketches solve the WSD problem, this type of mistakes is important for us.

\begin{figure}[H]
\centering
\includegraphics[width=15cm]{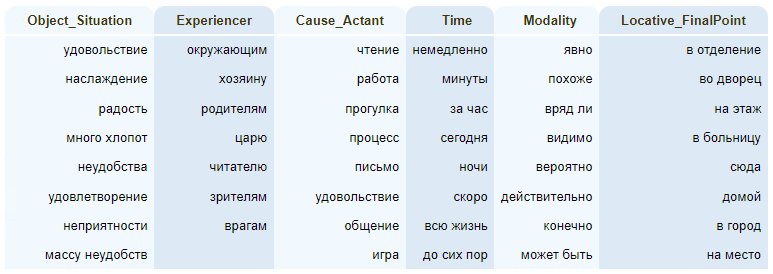}
\caption{The sketch for the verb ’\foreignlanguage{russian}{доставлять}:TO\_BRING\_STATE\_TO\_SMB’}  
\label{fig:deliver}
\end{figure}

An illustration for the incorrect semantic slot is the Russian sketch for ‘\foreignlanguage{russian}{доставлять}:TO\_BRING\_STATE\_TO\_SMB’ (fig. \ref{fig:deliver}), parallel for the above shown ‘inflict:TO\_BRING\_STATE\_TO\_SMB’. It contains the [Locative\_FinalPoint] slot, which must definitely belong to another meaning of the verb -- ‘bring to some place’.

Examples of the wrong fillers have already been shown in \cite{semsketches2021}.
The reasons are usually bound either with the statistics, or with the work of the parser. At the analysis stage, all possible hypotheses are built for the sentence -- with all possible homonyms that can fit. The final structure turns out to be the one with the highest scores. In some cases, hypotheses with more frequent homonyms win due to their higher frequency, in spite of the fact that the whole structure with the wrong homonym gets lower evaluations.

As the text collections for building the sketches grow, the statistics of the proper analysis improves, therefore, we expect that most part of the errors will be corrected with enlarging the corpora. Nevertheless, in case of the improper work of the parser, the opportunity to correct the semantic models that the parser uses exists as well. 

\subsection{The syntactic homonymy}

Key difference between the semantic and the syntactic sketches is that in the former 1 surface realisation can correspond to various semantic roles. For instance, ‘for’-dependency can introduce Time, Purpose, Distance, Motive and a number of other relations.

Usually, the proper semantic role is chosen according to the semantic model of the given verb in Compreno -- namely, the set of the semantic slots with the necessary surface realisation, the fillers of the semantic slots, and their status (which marks the role as more or less preferable).

\begin{figure}[H]
\centering
\includegraphics[width=15cm]{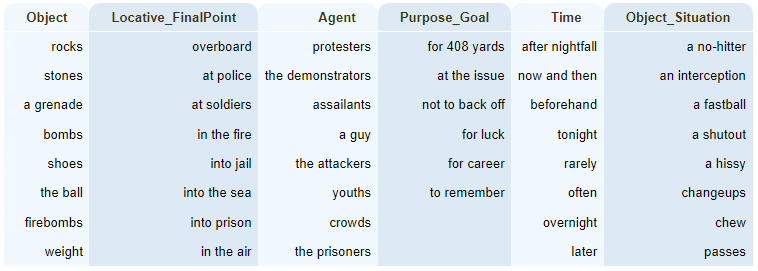}
\caption{The sketch for the verb ‘to throw:TO\_THROW’}  
\label{fig:throw}
\end{figure}

When the model or the statistics give improper results, the semantic role of the dependency can be defined incorrectly. For instance, see the [Purpose\_Goal] slot of the verb ‘throw:TO\_THROW’: the first line contains the nominal group ‘for 408 yards’, which must evidently belong to the [Locative\_Distance] slot (fig. \ref{fig:throw}).

\begin{figure}[H]
\centering
\includegraphics[width=15cm]{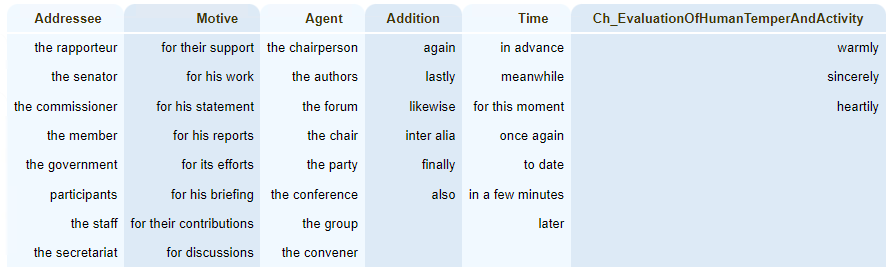}
\caption{The sketch for the verb ‘to thank:TO\_THANK’} 
\label{fig:thank}
\end{figure}

Another example is the group ‘for this moment’ in the [Time] slot instead of [Motive] in the sketch of ‘to thank’ (fig. \ref{fig:thank}). Here, on the contrary, [Motive] is definitely more frequent, but ‘moment’ is a very typical [Time] filler, therefore, high statistical evaluation of the correlation ‘moment’-[Time] made the incorrect structure win.

All the mistakes deal with different aspects of the WSD and homonymy problems. Their number does not seem significant, nevertheless, their statistical estimation must be made when creating a larger sketch corpus.

\section{Cross-language differences between the sketches with similar semantics}

The one-language sketch corpora suggest good lexicographic portraits of the verbs, showing their most frequent semantic links sorted according to the semantic roles of the dependencies. Moreover, apart from purely lexicographic tasks, the sketches allow one to solve various problems bound both with the context usage of the verbs and with their polysemy.

Another purpose of the sketches deals with contrastive studies. Parallel sketches from different languages give perfect representation of the correlation between similar verbs, therefore, parallel sketch corpora would be helpful in this respect.

Evidently, each sketch can correspond to more than one sketch in another language. To get a full set of all possible counterparts, one should take the necessary sketch in one language and the sketches for all the semantic equivalents in the same semantic class in another language. After it, one can range the counterparts according to their affinity with the primary verb. We have not made such full sets in the pilot corpus, however, adding this option is included in further plans.

At the current stage of the project, the correlations between the English and the Russian sketches do not include all possible correlations for each verb, so the sketch pairs are just a subset of the possible variants.

Some pairs look similar: both English and Russian sketches include the same set of semantic roles, and the semantic roles contain either fillers with close semantics, or just a wide range of fillers with no special semantic restrictions on them. 

At the same time, many sketches demonstrate significant differences between the English and Russian equivalents. Most of them concern the following situations:
\begin{enumerate}[label=(\alph*)]
\item some semantic slot is present in the sketch of one language and is absent -- in the corresponding sketch in another language;
\item equivalent sketches contain the same sets of the roles in both languages, but the fillers of some role differ significantly;
\item the semantic field where the considered verbs belong is structured differently in different languages.
\end{enumerate}

\subsection{Different semantic roles in the equivalent sketches from different languages}

Frequently, the semantic role sets in the parallel sketches do not coincide completely. It concerns both the actant roles and the circumstantial ones. The reasons can be different. First, the semantics of one verb may be wider than the semantics of the other, therefore, the model of the former can include additional roles which are absent in the model of the latter. Second, the model of both verbs can include the same sets of roles, however, the frequency of some roles may differ for various verbs, which can be motivated both by the verb’s semantics and by the representativeness and contents of the corpora for building the sketches. 

An example of the first case is the correlation between the semantic derivates in different languages. For instance, Russian verb ‘\foreignlanguage{russian}{трясти}’ ‘to shake’ does not attach the initial point dependency in contexts like (1) and (2), while the English ‘shake’ does:

\begin{enumerate}[label=(\arabic*)]
\item A sound they couldn’t shake [from their Locative\_InitialPoint: heads] -- \foreignlanguage{russian}{Звук, который им никак не удавалось вытряхнуть [из} Locative\_InitialPoint: \foreignlanguage{russian}{головы];}
\item I saw immediately that my few belongings had been disturbed–collars not refolded, one of my chemises balled up and pushed into a corner, the tortoiseshell comb shaken [from its Locative\_InitialPoint: handkerchief]. -- \foreignlanguage{russian}{И сразу увидела, что в моих вещах кто-то рылся — воротники были сложены неаккуратно, одна из моих рубашек скомкана и засунута в угол, черепаховый гребень вытряхнут [из носового} Locative\_InitialPoint: \foreignlanguage{russian}{платка]}.
\end{enumerate}

In Russian, the semantic derivate \russianexample{вытряхнуть}{shake out} is used when the initial point role is expressed in a sentence. Therefore, the sketches can show that ‘shake’ usually corresponds to the Russian ‘\foreignlanguage{russian}{трясти}’ (which does not mark the ‘direction of shaking’), but can also correspond to ‘\foreignlanguage{russian}{вытрясти}’ (which denotes the ‘from’ direction) with the dependency of the initial point.

Nevertheless, there can be occasional variations depending on the contents of the corpora, especially as far as less frequent verbs are considered. The more the corpora are, the more stable are the results. Thus we permanently enlarge the size of the dataset for building the sketches.

\begin{figure}[H]
\centering
\includegraphics[width=13cm]{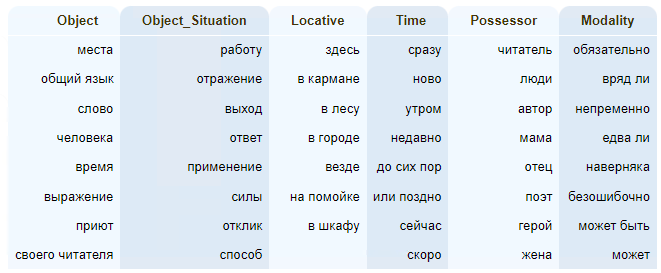}
\caption{The sketch for the verb ‘\foreignlanguage{russian}{найти}:TO\_SEEK\_FIND’} 
\label{fig:nayti}
\end{figure}

As an instance of such statistical oscillations, see the sketches for “find:TO\_SEEK\_FIND” and “\foreignlanguage{russian}{найти}:TO\_SEEK\_FIND”. The first five roles coincide, but the sixth one is different -- it is [Metaphoric\_Locative] for the English ‘find’ and [Modality] for the Russian ‘\foreignlanguage{russian}{найти}’ (fig. \ref{fig:find}, \ref{fig:nayti}):

\begin{figure}[H]
\centering
\includegraphics[width=13cm]{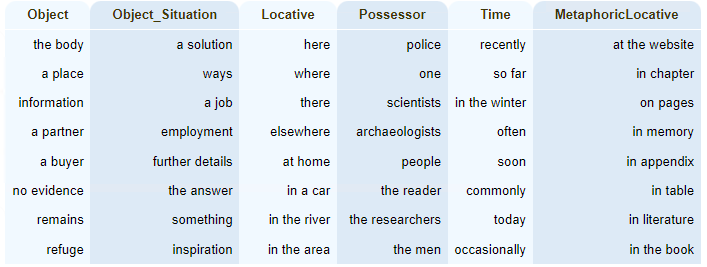}
\caption{The sketch for the verb ‘to find:TO\_SEEK\_FIND’} 
\label{fig:find}
\end{figure}

Both roles -- [Metaphoric\_Locative] and [Modality] -- can be frequently used with both verbs. In this case, the difference does not seem meaningful.

\subsection{Different fillers of the semantic roles}

Let us consider some sketches for the descendants of the semantic class “TO\_COMMIT”: the English verbs ‘do’, ‘play’ and the Russian verbs ‘\foreignlanguage{russian}{делать}’,‘\foreignlanguage{russian}{играть}’. “TO\_COMMIT” is a kind of lexical function, where the verbs have rather narrow compatibility in the [Object] role (place trust/hope vs pay a visit vs play a joke/trick vs take a look/try/walk/etc., and so on).

\begin{figure}[H]
\centering
\includegraphics[width=13cm]{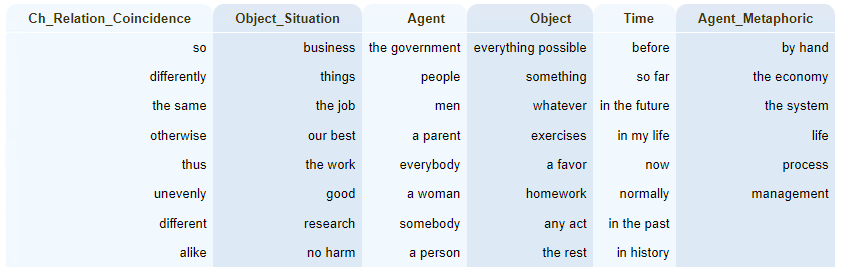}
\caption{The sketch for the verb ‘to do:TO\_COMMIT’}  
\label{fig:do}
\end{figure}

\begin{figure}[H]
\centering
\includegraphics[width=13cm]{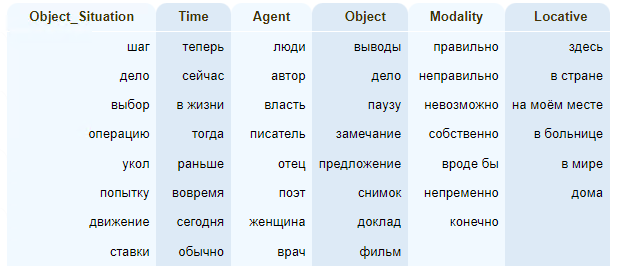}
\caption{The sketch for the verb ‘\foreignlanguage{russian}{делать}:TO\_COMMIT’}  
\label{fig:delat}
\end{figure}

\begin{figure}[H]
\centering
\includegraphics[width=13cm]{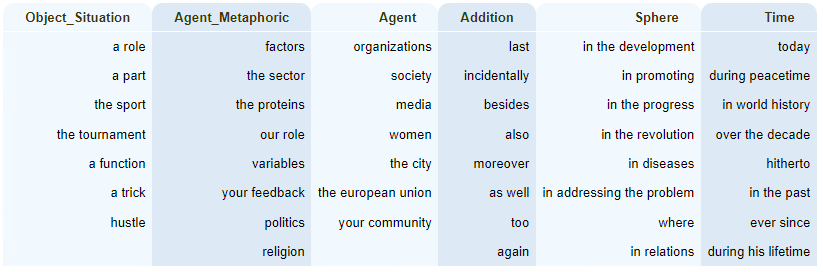}
\caption{The sketch for the verb ‘to play:TO\_COMMIT’}  
\label{fig:play}
\end{figure}

\begin{figure}[H]
\centering
\includegraphics[width=13cm]{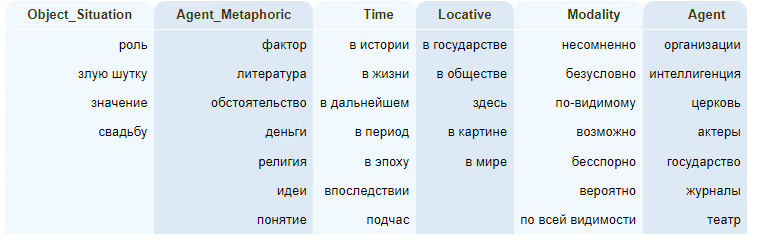}
\caption{The sketch for the verb ‘\foreignlanguage{russian}{играть}:TO\_COMMIT’}  
\label{fig:igrat}
\end{figure}

As fig. \ref{fig:do}, \ref{fig:delat}, \ref{fig:play} and \ref{fig:igrat} demonstrate, the compatibility of the verbs ‘do’ and ‘\foreignlanguage{russian}{делать}’ is rather wide, while ‘\foreignlanguage{russian}{играть}’ combines with only four Object fillers. 


Besides, the four verbs differ in the sets of the semantic roles as well. [Agent], [Object\_Situation] and [Time] are present in all four sketches. [Object] is absent in the sketches of ‘\foreignlanguage{russian}{играть}’ and ‘play’ as their compatibility does not include the corresponding fillers. 

‘Do’ and ‘\foreignlanguage{russian}{играть}/play’ include [Agent\_Metaphoric] slot, while ‘\foreignlanguage{russian}{делать}’ does not include it. The reason seems to be in the semantics of the fillers of the [Object] and [Object\_Situation] slots: the most frequent Object\_Class fillers are \russianexample{шаг}{step}, \russianexample{выбор}{choice}, \russianexample{операция}{operation}, \russianexample{снимок}{picture} and so on, which are more often combined with active human-like agent rather than inanimate agents like ‘economy, system, process’ and alike.

As far as the circumstantial dependencies are concerned, both Russian sketches include the semantic roles of [Modality] and [Locative] while the English ‘do’ includes [Ch\_Relation\_Coincidence] slot (in the Compreno model, it characterizes objects or situations according to their similarity) and ‘play’ -- [Addition] and [Sphere]. At first sight, these differences do not seem meaningful, however, it would be interesting to regard the sketches of the whole semantic class TO\_COMMIT to examine how regular such correlations are.

\begin{figure}[H]
\includegraphics[width=15cm]{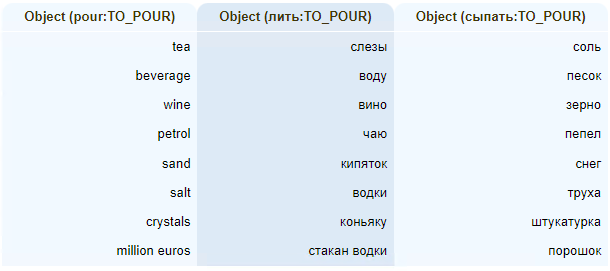}
\caption{The fragment of the sketches for the verbs ‘to pour:TO\_POUR’, ‘\foreignlanguage{russian}{лить}:TO\_POUR, and ‘\foreignlanguage{russian}{сыпать}:TO\_POUR’}  
\label{fig:pour}
\end{figure}

Another example concerns verbs with wider compatibility, where the restrictions on the Object role are not purely lexicalized, but concern a wider range of fillers with common semantic features. For instance, let us take the semantic field “TO\_POUR” (something liquid or friable). English and Russian structure it differently as far as the core verbs’ compatibility is concerned. Namely, the English verb ‘to pour’ attaches objects which are liquid (water, wine), friable (sand, sugar), or consist of many small pieces (crystals, euros, diced meat, and so on). In Russian, the verb ‘\foreignlanguage{russian}{лить}’ is used with liquid objects only and the verb ‘\foreignlanguage{russian}{сыпать}’ -- only with friable objects and objects consisting of many small pieces. Therefore, the Object slot fillers differ correspondingly in the sketches (fig. \ref{fig:pour}).

Nonetheless, the amount of eight most frequent fillers which is usually shown in the sketches is not always enough to demonstrate such differences, as the most frequent objects can bear the same semantic features.

As one can see, the sketches suggest a wide range of comparative data in the field of semantics and demonstrate the semantical differences between the verbs of the same semantic class both in different languages and within one language as well. 

\section{Conclusion}

In the given paper, we have presented the pilot corpus of the English semantic sketches. 

As the sketches are provided with their semantic parallels in Russian, we have also illustrated what kind of comparative studies the sketches allow to conduct, especially as far as the differences in the semantic roles and their typical fillers are concerned. An important point is the ability of the sketches to deal with polysemy and to differentiate between various homonyms. 

We have also discussed common types of mistakes occurring while building the sketches and speculated about their linguistic and technical nature. 

Our further plans are to improve the sketches by obtaining them on a bigger dataset, to enlarge the sketch corpus and build the sketches for each verb from the dataset, to provide the corpus with some additional features, such as the opportunity to show more semantic slots and more fillers of the slots when necessary, and to see the correlations between all the verbs of the same semantic class.
After it, the work on adding other languages to the sketch corpus will be started.

At the same time, we work on the open corpus of the Compreno semantic mark-up which will include a detailed description of the mark-up principles and the semantic roles used in the mark-up, which will facilitate the understanding of the roles used in the sketches. 

The current corpus is available at github\footnote{https://github.com/dialogue-evaluation/SemSketches/tree/main/data/task\_2}. Besides, we continue the  work on integrating the semantic sketches in the General Internet-Corpus of Russian (GICR).

We hope the corpus would contribute to different NLP areas, especially to solving the WSD problem.

\bibliography{dialogue.bib}
\bibliographystyle{dialogue}



\end{document}